# Reasoning about Expectation

**Joseph Y. Halpern**
Department of Computer Science
Cornell University
Ithaca, NY 14853
halpern@cs.cornell.edu
http://www.cs.cornell.edu/home/halpern

**Riccardo Pucella**
Department of Computer Science
Cornell University
Ithaca, NY 14853
riccardo@cs.cornell.edu

## Abstract

Expectation is a central notion in probability theory. The notion of expectation also makes sense for other notions of uncertainty. We introduce a propositional logic for reasoning about expectation, where the semantics depends on the underlying representation of uncertainty. We give sound and complete axiomatizations for the logic in the case that the underlying representation is (a) probability, (b) sets of probability measures, (c) belief functions, and (d) possibility measures. We show that this logic is more expressive than the corresponding logic for reasoning about likelihood in the case of sets of probability measures, but equi-expressive in the case of probability, belief, and possibility. Finally, we show that satisfiability for these logics is NP-complete, no harder than satisfiability for propositional logic.

## 1 INTRODUCTION

One of the most important notions in probability theory is that of *expectation*. The expected value of a random variable is, in a sense, the single number that best describes the random variable. In this paper, we consider the notion of expectation in a more general setting.

It is well known that a probability measure determines a unique expectation function that is linear (i.e., $E(aX + bY) = aE(X) + bE(Y)$), monotone (i.e., $X \leq Y$ implies $E(X) \leq E(Y)$), and maps constant functions to their value. Conversely, given an expectation function $E$ (that is, a function from random variables to the reals) that is linear, monotone, and maps constant functions to their value, there is a unique probability measure $\mu$ such that $E = E_\mu$. That is, there is a 1-1 mapping from probability measures to (probabilistic) expectation functions.

Walley [1991] has argued persuasively that it is necessary to occasionally go beyond probabilistic expectation. He introduces lower and upper *previsions*, which are essentially lower and upper expectations with respect to sets of probability measures. Characterizations of the epectation functions that arise from sets of probability measures are well known [Huber 1981; Walley 1981; Walley 1991]. However, there seems to be surprisingly little work on characterizing expectation in the context of other measures of uncertainty, such as belief functions [Shafer 1976] and possibility measures [Dubois and Prade 1990]. We provide characterizations here.

Having characterized expectation functions, we then turn to the problem of reasoning about them. We define a logic similar in spirit to that introduced in [Fagin, Halpern, and Megiddo 1990] (FHM from now on) for reasoning about likelihood expressed as either probability or belief. The same logic is used in [Halpern and Pucella 2001] (HP from now on) for reasoning about upper probabilities. The logic for reasoning about expectation is strictly more expressive than its counterpart for reasoning about likelihood if the underlying semantics is given in terms of sets of probability measures (so that upper probabilities and upper expectations are used, respectively); it turns out to be equi-expressive in the case of probability, belief functions, and possibility measures. This is somewhat surprising, especially in the case of belief functions. In all cases, the fact that expectations are at least as expressive is immediate, since the expectation of $\varphi$ (viewed as an indicator function, that is, the random variable that is 1 in worlds where $\varphi$ is true and 0 otherwise) is equal to its likelihood. However, it is not always obvious how to express the expectation of a random variable in terms of likelihood.

We then provide a sound and complete axiomatization for the logic with respect to each of the interpretations of expectation that we consider, using our characterization of expectation. Finally, we show that, just as in the case of the corresponding logic for reasoning about likelihood, the complexity of the satisfiability problem is NP-complete. This is clear when the underlying semantics is given in terms of probability measures, belief functions, or possibility measures, but it is perhaps surprising that, despite the added expressiveness in the case of sets of probability measures, reasoning in the logic remains NP-complete.

To the best of our knowledge, there is only one previous attempt to express properties of expectation in a logical setting. Wilson and Moral [1994] takes as their starting point Walley's notion of lower and upper previsions. They consider when acceptance of one set of gambles implies acceptance of another gamble. This is a notion that is easily expressible in our logic when the original set of gambles is finite, so our logic subsumes theirs in the finite case.

This paper is organized as follows. In the next section, the characterizations of expectation for probability measures and sets of probability measures are reviewed, and the characterizations of expectation for belief functions and possibility measures are provided. In Section 3, we introduce a logic for reasoning about expectation with respect to all these representations of uncertainty. In Section 4, we compare the expressive power of our expectation logic to that of the logic for reasoning about likelihood. In Section 5, we derive sound and complete axiomatizations for the logic in Section 3, with respect to different representations of uncertainty. In Section 6, we discuss an axiomatization of gamble inequalities, which is assumed by the axiomatizations given in Section 5. Finally, in Section 7, we prove that the decision problem for the expectation logic is NP-complete for each of the representations of uncertainty we consider. Due to lack of space, proofs are left to the full paper.

## 2 EXPECTATION FUNCTIONS

Recall that a *random variable* $X$ on a sample space (set of possible worlds) $W$ is a function from $W$ to some range. Let $\mathcal{V}(X)$ denote the image of $X$, that is, the possible values of $X$. A *gamble* is a random variable whose range is the reals. In this paper, we focus on the expectation of gambles; we consider only gambles $X$ such that $\mathcal{V}(X)$ is finite. (This allows us to define expectation using summation rather than integration.)

### 2.1 EXPECTATION FOR PROBABILITY MEASURES

Given a probability measure $\mu$ and a gamble $X$, the *expected value of $X$* (or the *expectation of $X$*) with respect to $\mu$, denoted $E_\mu(X)$, is just

$$E_\mu(X) = \sum_{x \in \mathcal{V}(X)} x\mu(X = x). \tag{1}$$

This definition implicitly assumes that the gamble $X$ is *measurable*, that is, for each value $x \in \mathcal{V}(X)$, the set of worlds $X = x$ where $X$ takes on value $x$ is measurable.

As is well known, probabilistic expectation functions can be characterized by a small collection of properties. If $X$ and $Y$ are gambles on $W$ and $a$ and $b$ are real numbers, define the gamble $aX + bY$ on $W$ in the obvious way: $(aX + bY)(w) = aX(w) + bY(w)$. Say that $X \leq Y$ if $X(w) \leq Y(w)$ for all $w \in W$. Let $\tilde{c}$ denote the constant function which always returns $c$; that is, $\tilde{c}(w) = c$. Let $\mu$ be a probability measure on $W$.

**Proposition 2.1:** *The function $E_\mu$ has the following properties for all measurable gambles $X$ and $Y$.*

(a) $E_\mu$ *is additive:* $E_\mu(X + Y) = E_\mu(X) + E_\mu(Y)$.

(b) $E_\mu$ *is affinely homogeneous:* $E_\mu(aX + \tilde{b}) = aE_\mu(X) + b$ *for all $a, b \in \mathbb{R}$.*

(c) $E_\mu$ *is monotone:* *if $X \leq Y$, then $E_\mu(X) \leq E_\mu(Y)$.*

The next result shows that the properties in Proposition 2.1 essentially characterize probabilistic expectation functions. It too is well known.

**Theorem 2.2:** *Suppose that $E$ maps gambles measurable with respect to some algebra $\mathcal{F}$ to $\mathbb{R}$ and $E$ is additive, affinely homogeneous, and monotone. Then there is a (necessarily unique) probability measure $\mu$ on $\mathcal{F}$ such that $E = E_\mu$.*

### 2.2 EXPECTATION FOR SETS OF PROBABILITY MEASURES

If $\mathcal{P}$ is a set of probability measures on a space $W$, define

$$\mathcal{P}_*(U) = \inf\{\mu(U) : \mu \in \mathcal{P}\} \text{ and}$$
$$\mathcal{P}^*(U) = \sup\{\mu(U) : \mu \in \mathcal{P}\}.$$

$\mathcal{P}_*(U)$ is called the *lower probability* of $U$ and $\mathcal{P}^*(U)$ is called the *upper probability* of $U$. Lower and upper probabilities have been well studied in the literature (see, for example, [Borel 1943; Smith 1961]).

There are straightforward analogues of lower and upper probability in the context of expectation. If $\mathcal{P}$ is a set of probability measures such that $X$ is measurable with respect to each probability measure $\mu \in \mathcal{P}$, then define $E_\mathcal{P}(X) = \{E_\mu(X) : \mu \in \mathcal{P}\}$. $E_\mathcal{P}(X)$ is a set of numbers. Define the *lower expectation* and *upper expectation* of $X$ with respect to $\mathcal{P}$, denoted $\underline{E}_\mathcal{P}(X)$ and $\overline{E}_\mathcal{P}(X)$, as the inf and sup of the set $E_\mathcal{P}(X)$, respectively.

The properties of $\underline{E}_\mathcal{P}$ and $\overline{E}_\mathcal{P}$ are not so different from those of probabilistic expectation functions. Let $X_U$ denote the gamble such that $X_U(w) = 1$ if $w \in U$ and $X_U(w) = 0$ if $w \notin U$. A gamble of the form $X_U$ is traditionally called an *indicator function*. Note that $E_\mu(X_U) = \mu(U)$. Similarly, it is easy to see that $\mathcal{P}_*(U) = \underline{E}_\mathcal{P}(X_U)$ and $\mathcal{P}^*(U) = \overline{E}_\mathcal{P}(X_U)$. Moreover, we have the following analogue of Propositions 2.1.

**Proposition 2.3:** *The functions $\underline{E}_\mathcal{P}$ and $\overline{E}_\mathcal{P}$ have the following properties for all gambles $X$ and $Y$.*

(a) $\underline{E}_\mathcal{P}(X+Y) \geq \underline{E}_\mathcal{P}(X) + \underline{E}_\mathcal{P}(Y)$ *(superadditivity)*;
$\overline{E}_\mathcal{P}(X+Y) \leq \overline{E}_\mathcal{P}(X) + \overline{E}_\mathcal{P}(Y)$ *(subadditivity)*.

(b) $\underline{E}_\mathcal{P}$ *and* $\overline{E}_\mathcal{P}$ *are both* positively affinely homogeneous: $\underline{E}_\mathcal{P}(aX+b) = a\underline{E}_\mathcal{P}(X) + b$ *and* $\overline{E}_\mathcal{P}(aX+b) = a\overline{E}_\mathcal{P}(X) + b$ *if* $a, b \in \mathbb{R}, a \geq 0$.

(c) $\underline{E}_\mathcal{P}$ *and* $\overline{E}_\mathcal{P}$ *are monotone.*

(d) $\overline{E}_\mathcal{P}(X) = -\underline{E}_\mathcal{P}(-X)$.

Superadditivity (resp., subadditivity), positive affine homogeneity, and monotonicity in fact characterize $\underline{E}_\mathcal{P}$ (resp., $\overline{E}_\mathcal{P}$).

**Theorem 2.4:** [Huber 1981] *Suppose that $E$ maps gambles measurable with respect to $\mathcal{F}$ to $\mathbb{R}$ and is superadditive (resp., subadditive), positively affinely homogeneous, and monotone. Then there is a set $\mathcal{P}$ of probability measures on $\mathcal{F}$ such that $E = \underline{E}_\mathcal{P}$ (resp., $E = \overline{E}_\mathcal{P}$).*[1]

The set $\mathcal{P}$ constructed in Theorem 2.4 is not unique. It is not hard to construct sets $\mathcal{P}$ and $\mathcal{P}'$ such that $\mathcal{P} \neq \mathcal{P}'$ but $\underline{E}_\mathcal{P} = \underline{E}_{\mathcal{P}'}$. However, there is a canonical largest set $\mathcal{P}$ such that $E = \underline{E}_\mathcal{P}$; $\mathcal{P}$ consists of all probability measures $\mu$ such that $E_\mu(X) \geq E(X)$ for all gambles $X$. This set $\mathcal{P}$ can be shown to be closed and convex. Indeed, it easily follows that Theorem 2.4 actually provides a 1-1 mapping from closed, convex sets of probability measures to lower/upper expectations. Moreover, in a precise sense, this is the best we can do. If $\mathcal{P}$ and $\mathcal{P}'$ have the same convex closure (where the *convex closure* of a set is the smallest closed, convex set containing it), then $\underline{E}_\mathcal{P} = \underline{E}_{\mathcal{P}'}$.

As Walley [1991] shows, what he calls *coherent lower/upper previsions* are also lower/upper expectations with respect to some set of probability measures. Thus, lower/upper previsions can be identified with closed, convex sets of probability measures.

### 2.3 EXPECTATION FOR BELIEF FUNCTIONS

As is well known, a belief function [Shafer 1976] Bel is a function from subsets of a state space $W$ to $[0, 1]$ satisfying the following three properties:

B1. $\mathrm{Bel}(\emptyset) = 0$.

B2. $\mathrm{Bel}(W) = 1$.

B3. For $n = 1, 2, 3, \ldots$, $\mathrm{Bel}(\bigcup_{i=1}^n U_i) \geq$
$\sum_{i=1}^n \sum_{\{I \subseteq \{1,\ldots,n\}:|I|=i\}} (-1)^{i+1} \mathrm{Bel}(\bigcap_{j \in I} U_j)$.

Given a belief function Bel, there is a corresponding *plausibility function* Plaus, where $\mathrm{Plaus}(U) = 1 - \mathrm{Bel}(\overline{U})$. It follows easily from B3 that $\mathrm{Bel}(U) \leq \mathrm{Plaus}(U)$ for all $U \subseteq W$. $\mathrm{Bel}(U)$ can be thought of as a lower bound of a set of probabilities and $\mathrm{Plaus}(U)$ can be thought of as the corresponding upper bound. This intuition is made precise in the following well-known result.

**Theorem 2.5:** [Dempster 1967] *Given a belief function* Bel *defined on $W$, let $\mathcal{P}_{\mathrm{Bel}} = \{\mu : \mu(U) \geq \mathrm{Bel}(U)$ for all $U \subseteq W\}$. Then* $\mathrm{Bel} = (\mathcal{P}_{\mathrm{Bel}})_*$.

There is an obvious way to define a notion of expectation based on belief functions, using the identification of Bel with $(\mathcal{P}_{\mathrm{Bel}})_*$. Given a belief function Bel, define $E_{\mathrm{Bel}} = \underline{E}_{\mathcal{P}_{\mathrm{Bel}}}$. Similarly, for the corresponding plausibility function Plaus, define $E_{\mathrm{Plaus}} = \overline{E}_{\mathcal{P}_{\mathrm{Bel}}}$. (These definitions are in fact used by Dempster [1967]).

This is well defined, but it seems more natural to get a notion of expectation for belief functions that is defined purely in terms of belief functions, without reverting to probability. One way of doing so is due to Choquet [1953].[2] It takes as its point of departure the following alternate definition of expectation in the case of probability. Suppose $X$ is a gamble such that $\mathcal{V}(X) = \{x_1, \ldots, x_n\}$, with $x_1 < \ldots < x_n$.

**Proposition 2.6:** $E_\mu(X) = x_1 + (x_2 - x_1)\mu(X > x_1) + \cdots + (x_n - x_{n-1})\mu(X > x_{n-1})$.

Define

$$E'_{\mathrm{Bel}}(X) = x_1 + (x_2 - x_1)\mathrm{Bel}(X > x_1) + \cdots + (x_n - x_{n-1})\mathrm{Bel}(X > x_{n-1}). \quad (2)$$

An analogous definition holds for plausibility:

$$E'_{\mathrm{Plaus}}(X) = x_1 + (x_2 - x_1)\mathrm{Plaus}(X > x_1) + \cdots + (x_n - x_{n-1})\mathrm{Plaus}(X > x_{n-1}). \quad (3)$$

**Proposition 2.7:** [Schmeidler 1989] $E_{\mathrm{Bel}} = E'_{\mathrm{Bel}}$ *and* $E_{\mathrm{Plaus}} = E'_{\mathrm{Plaus}}$.

Thus, using (2), there is a way of defining expectation for belief functions without referring to probability.

Since $E_{\mathrm{Bel}}$ can be viewed as a special case of the lower expectation $\underline{E}_\mathcal{P}$ (taking $\mathcal{P} = \mathcal{P}_{\mathrm{Bel}}$), it is immediate from Proposition 2.3 that $E_{\mathrm{Bel}}$ is superadditive, positively affinely homogeneous, and monotone. (Similar remarks hold for $E_{\text{Plaus}}$, except that it is subadditive. For ease of exposition, we focus on $E_{\text{Bel}}$ in the remainder of this section, although analogous remarks hold for $E_{\text{Plaus}}$.) But $E_{\text{Bel}}$ has additional properties.

---

[1] There is an equivalent characterization of $\underline{E}_\mathcal{P}$, due to Walley [1991]. He shows that $E = \underline{E}_\mathcal{P}$ for some set $\mathcal{P}$ of probability measures iff $E$ is superadditive, $E(cX) = cE(X)$, and $E(X) \geq \inf\{X(w) : w \in W\}$. An analogous characterization holds for $\overline{E}_\mathcal{P}$.

[2] Choquet actually talked about $k$-monotone capacities, which are essentially functions that satisfy B3 where $n = 1, \ldots, k$. Belief functions are essentially equivalent to infinitely monotone capacities.

Since it is immediate from the definition that $E_{\text{Bel}}(X_U) = \text{Bel}(U)$, the inclusion-exclusion property B3 of belief functions can be expressed in terms of expectation (just by replacing all instances of $\text{Bel}(V)$ in B3 by $E_{\text{Bel}}(X_V)$). Moreover, it does not follow from the other properties, since it can be shown not to hold for arbitrary lower probabilities. B3 seems like a rather specialized property, since it applies only to indicator functions. There is a more general version of it that also holds for $E_{\text{Bel}}$. Given gambles $X$ and $Y$, define the gambles $X \wedge Y$ and $X \vee Y$ as the minimum and maximum of $X$ and $Y$, respectively; that is, $(X \wedge Y)(w) = \min(X(w), Y(w))$ and $(X \vee Y)(w) = \max(X(w), Y(w))$. Consider the following inclusion-exclusion rule for expectation functions:

$$E(\vee_{i=1}^n X_i) \geq \sum_{i=1}^n \sum_{\{I \subseteq \{1,\ldots,n\}: |I|=i\}} (-1)^{i+1} E(\wedge_{j \in I} X_j). \quad (4)$$

Since it is immediate that $X_{U \cup V} = X_U \vee X_V$ and $X_{U \cap V} = X_U \wedge X_V$, (4) generalizes B3.

There is yet another property satisfied by expectation functions based on belief functions. It too is expressed in terms of indicator functions; we do not know if there is a generalization that holds for arbitrary gambles.

If $U_1 \supseteq U_2 \supseteq \ldots \supseteq U_n$ and $a_1, \ldots, a_n \geq 0$
then $E(a_1 X_{U_1} + \cdots + a_n X_{U_n}) = a_1 E(X_{U_1}) + \cdots + a_n E(X_{U_n}).$ (5)

**Proposition 2.8:** *The function $E_{\text{Bel}}$ is superadditive, positively affinely homogeneous, monotone, and satisfies (4) and (5).*

**Theorem 2.9:** *Suppose that $E$ is an expectation function that is positively affinely homogeneous, monotone, and satisfies (4) and (5). Then there is a (necessarily unique) belief function $\text{Bel}$ such that $E = E_{\text{Bel}}$.*

Note that superadditivity was not assumed in the statement of Theorem 2.9. This suggests that superadditivity follows from the other properties. This is indeed the case.

**Proposition 2.10:** *An expectation function that satisfies (4) and (5) is superadditive.*

It follows easily from these results that $E_{\text{Bel}}$ is the unique expectation function $E$ that is superadditive, positively affinely homogeneous, monotone, and satisfies (4) and (5) such that $E(X_U) = \text{Bel}(U)$ for all $U \subseteq W$. Proposition 2.8 shows that $E_{\text{Bel}}$ has these properties. If $E'$ is an expectation function that has these properties, by Theorem 2.9, $E' = E_{\text{Bel}'}$ for some belief function $\text{Bel}'$.

Since $E'(X_U) = \text{Bel}'(U) = \text{Bel}(U)$ for all $U \subseteq W$, it follows that $\text{Bel} = \text{Bel}'$.

This is somewhat surprising. While it is almost immediate that an additive, affinely homogeneous expectation function (the type that arises from a probability measure) is determined by its behavior on indicator functions, it is not at all obvious that a superadditive, positively affine homogeneous expectation function should be determined by its behavior on indicator functions. In fact, in general it is not, as the following example shows. The inclusion-exclusion property is essential. This observation says that $\text{Bel}$ and $E_{\text{Bel}}$ contain the same information. Thus, so do $(\mathcal{P}_{\text{Bel}})_*$ and $\underline{E}_{\mathcal{P}_{\text{Bel}}}$ (since $\text{Bel} = (\mathcal{P}_{\text{Bel}})_*$ and $E_{\text{Bel}} = \underline{E}_{\mathcal{P}_{\text{Bel}}}$). However, this is not true for arbitrary sets $\mathcal{P}$ of probability measures, as the following example shows. Let $W = \{1, 2, 3\}$. A probability measure $\mu$ on $W$ can be characterized by a triple $(a_1, a_2, a_3)$, where $\mu(i) = a_i$. Let $\mathcal{P}$ consist of the three probability measures $(0, 3/8, 5/8)$, $(5/8, 0, 3/8)$, and $(3/8, 5/8, 0)$. It is almost immediate that $\mathcal{P}_*$ is 0 on singleton subsets of $W$ and $\mathcal{P}_* = 3/8$ for doubleton subsets. Let $\mathcal{P}' = \mathcal{P} \cup \{\mu_4\}$, where $\mu_4 = (5/8, 3/8, 0)$. It is easy to check that $\mathcal{P}'_* = \mathcal{P}_*$. However, $\underline{E}_{\mathcal{P}} \neq \underline{E}_{\mathcal{P}'}$. In particular, let $X$ be the gamble such that $X(1) = 1$, $X(2) = 2$, and $X(3) = 3$. Then $\underline{E}_{\mathcal{P}}(X) = 13/8$ but $\underline{E}_{\mathcal{P}'}(X) = 11/8$. Thus, although $\underline{E}_{\mathcal{P}}$ and $\underline{E}_{\mathcal{P}'}$ agree on indicator functions, they do not agree on all gambles. In light of the discussion above, it should be no surprise that $\mathcal{P}_*$ is not a belief function.

### 2.4 EXPECTATION FOR POSSIBILITY MEASURES

A *possibility measure* Poss is a function from subsets of $W$ to $[0, 1]$ such that

Poss1. $\text{Poss}(\emptyset) = 0$.

Poss2. $\text{Poss}(W) = 1$.

Poss3. $\text{Poss}(\cup_i U_i) = \sup_i \text{Poss}(U_i)$.

It is well known [Dubois and Prade 1982] that a possibility measures are special cases of plausibility functions. Thus, (3) can be used to define a notion of possibilistic expectation; indeed, this has been done in the literature [Dubois and Prade 1987]. It is immediate from Poss3 that the expectation function $E_{\text{Poss}}$ defined from a possibility measure Poss in this way satisfies the *sup property*:

$$E_{\text{Poss}}(X_{\cup_i U_i}) = \max(E_{\text{Poss}}(X_{U_i})). \quad (6)$$

**Proposition 2.11:** *The function $E_{\text{Poss}}$ is positively affinely homogeneous, monotone, and satisfies (5) and (6).*

**Theorem 2.12:** *Suppose that $E$ is an expectation function that is positively affinely homogeneous, monotone, and*

*satisfies (5) and (6). Then there is a (necessarily unique) possibility measure* Poss *such that* $E = E_{\text{Poss}}$.

Note that, although Poss is a plausibility measure, and thus satisfies the analogue of (4) with $\geq$ replaced by $\leq$, there is no need to state (4) explicitly; it follows from (6). Moreover, just as in Lemma 2.10, it follows from the other properties that $E_{\text{Poss}}$ is subadditive. (Since a possibility measure is a plausibility function, not a belief function, the corresponding expectation function is subadditive rather than superadditive.)

## 3 A LOGIC FOR REASONING ABOUT EXPECTATION

We now consider a logic for reasoning about expectation. To set the stage, we briefly review the FHM logic for reasoning about likelihood.

### 3.1 REASONING ABOUT LIKELIHOOD

The syntax of the FHM logic is straightforward. Fix a set $\Phi_0 = \{p_1, p_2, \ldots\}$ of *primitive propositions*. The set $\Phi$ of *propositional formulas* is the closure of $\Phi_0$ under $\wedge$ and $\neg$. We assume a special propositional formula *true*, and abbreviate $\neg true$ as *false*. A *basic likelihood formula* has the form $a_1 \ell(\varphi_1) + \cdots + a_k \ell(\varphi_k) \geq b$, where $a_1, \ldots, a_k, b$ are real numbers and $\varphi_1, \ldots, \varphi_k$ are propositional formulas. The $\ell$ stands for *likelihood*. Thus, a basic likelihood formula talks about a linear combination of likelihood terms of the form $\ell_i(\varphi)$. A *likelihood formula* is a Boolean combination of basic likelihood formulas. Let $\mathcal{L}^{QU}$ be the language consisting of likelihood formulas. (The QU stands for *quantitative uncertainty*. The name for the logic is taken from [Halpern 2002].)

Although a likelihood formula involves $\geq$, it is easy to define similar expressions with $\leq$, $=$, $>$, and $<$, using the logical operators. For example, $a_1 \ell(\varphi_1) + \cdots + a_k \ell(\varphi_k) \leq b$ is an abbreviation for $-a_1 \ell(\varphi_1) - \cdots - a_k \ell(\varphi_k) \geq -b$. Using $\leq$ and $\geq$ gives us $=$; then $>$ can be obtained from $\geq$ and $=$, using negation.

The semantics of $\mathcal{L}^{QU}$ depends on how $\ell$ is interpreted. In FHM, it is interpreted as a probability measure and as a belief function; in HP, it is interpreted as an upper probability (determined by a set of probability measures). Thus, $\ell(\varphi)$ is taken as the probability of $\varphi$ (i.e., more precisely, the probability of the set of worlds where $\varphi$ is true), the belief in $\varphi$, etc. For example, in the case of probability, define a *probability structure* to be a tuple $M = (W, \mu, \pi)$, where $\mu$ is a probability on $W^3$ and $\pi$ is an *interpretation*, which associates with each state (or world) in $W$ a truth assignment on the primitive propositions in $\Phi_0$. Thus, $\pi(s)(p) \in \{\text{true}, \text{false}\}$ for $s \in W$ and $p \in \Phi_0$. Extend $\pi(s)$ to a truth assignment on all propositional formulas in the standard way, and associate with each propositional formula the set $[\![\varphi]\!]_M = \{s \in W : \pi(s)(\varphi) = \text{true}\}$. Then

$$M \models a_1 \ell(\varphi_1) + \cdots + a_n \ell(\varphi_n) \geq b \text{ iff}$$
$$a_1 \mu([\![\varphi_1]\!]_M) + \cdots + a_n \mu([\![\varphi_n]\!]_M) \geq b.$$

The semantics of Boolean combinations of basic likelihood formulas is given in the obvious way.

We can similarly give semantics to $\ell$ using lower (or upper) probability. Define a *lower probability structure* to be a tuple $M = (W, \mathcal{P}, \pi)$, $W$ and $\pi$ are, as before, a set of worlds and an interpretation, and $\mathcal{P}$ is a set of probability measures. Likelihood is interpreted as lower probability in lower probability structures:[4]

$$M \models a_1 \ell(\varphi_1) + \cdots + a_n \ell(\varphi_n) \geq b \text{ iff}$$
$$a_1 \mathcal{P}_*([\![\varphi_1]\!]_M) + \cdots + a_n \mathcal{P}_*([\![\varphi_n]\!]_M) \geq b.$$

A *belief structure* has the form $M = (W, \text{Bel}, \pi)$, where Bel is a belief function. We can interpret likelihood formulas with respect to belief structures in the obvious way. Similarly, a *possibility structure* has the form $M = (W, \text{Poss}, \pi)$, where Poss is a possibility measure. Again, we interpret likelihood formulas with respect to possibility structures in the obvious way.

Let $\mathcal{M}^{prob}$, $\mathcal{M}^{lp}$, $\mathcal{M}^{bel}$, and $\mathcal{M}^{poss}$ denote the set of all probability structures, lower probability structures, belief structures, and possibility structures, respectively.

### 3.2 REASONING ABOUT EXPECTATION

Our logic for reasoning about expectation is similar in spirit to $\mathcal{L}^{QU}$. The idea is to interpret a propositional formula $\varphi$ as the indicator function $X_{[\![\varphi]\!]_M}$, which is 1 in worlds where $\varphi$ is true, and 0 otherwise. We can then take linear combinations of such gambles. A *(linear) propositional gamble* is a linear combination of propositional formulas, of the form $b_1 \varphi_1 + \cdots + b_n \varphi_n$, where $b_1, \ldots, b_n$ are reals. We use $\gamma$ to represent propositional gambles. An *expectation inequality* is a statement of the form $a_1 e(\gamma_1) + \cdots + a_k e(\gamma_k) \geq b$, where $a_1 \ldots, a_k$ are reals, $k \geq 1$, and $b$ is a real. An *expectation formula* is a Boolean combination of expectation inequalities. We use $f$ and $g$ to represent expectation formulas. We use obvious abbreviations where needed, such as $e(\gamma_1) - e(\gamma_2) \geq a$ for $e(\gamma_1) + (-1)e(\gamma_2) \geq a$, $e(\gamma_1) \geq e(\gamma_2)$ for $e(\gamma_1) - e(\gamma_2) \geq 0$, $e(\gamma) \leq a$ for $-e(\gamma) \geq -a$, $e(\gamma) < a$ for $\neg(e(\gamma) \geq a)$ and $e(\gamma) = a$ for $(e(\gamma) \geq a) \wedge (e(\gamma) \leq a)$. Let $\mathcal{L}^E$ be the language consisting of expectation formulas.

---

[3] For simplicity here, we assume that all sets are measurable.

[4] In HP, we interpreted likelihood as upper probability. We interpret it here as lower probability to bring out the connections to belief, which is an instance of lower probability. It is easy to translate from upper probabilities to lower probabilities and vice versa, since $\mathcal{P}_*(U) = 1 - \mathcal{P}^*(\overline{U})$.

Given a model $M$, we associate with a propositional gamble $\gamma$ the gamble $\{\!|\gamma|\!\}_M$, where $\{\!|b_1\varphi_1 + \cdots + b_n\varphi_n|\!\}_M = b_1 X_{[\![\varphi_1]\!]_M} + \cdots + b_n X_{[\![\varphi_n]\!]_M}$. Of course, the intention is to interpret $e(\gamma)$ in $M$ as the expected value of the gamble $\{\!|\gamma|\!\}_M$, where the notion of "expected value" depends on the underlying semantics. In the case of probability structures, it is probabilistic expectation; in the case of belief structures, it is expected belief; in the case of lower probability structures, it is lower expectation; and so on. For example, if $M \in \mathcal{M}^{prob}$, then

$$M \models a_1 e(\gamma_1) + \cdots + a_k e(\gamma_k) \geq b \text{ iff}$$
$$a_1 E_\mu(\{\!|\gamma_1|\!\}_M) + \cdots + a_k E_\mu(\{\!|\gamma_k|\!\}_M) \geq b.$$

Again, Boolean combinations are defined in the obvious way. We leave the obvious semantic definitions in the case of belief structures and lower probability structures to the reader.

## 4 EXPRESSIVE POWER

It is easy to see that $\mathcal{L}^E$ is at least as expressive as $\mathcal{L}^{QU}$. Since the expected value of an indicator function is its likelihood, for all the notions of likelihood we are considering, replacing all occurrences of $\ell(\varphi)$ in a formula in $\mathcal{L}^{QU}$ by $e(\varphi)$ gives an equivalent formula in $\mathcal{L}^E$. Is $\mathcal{L}^E$ strictly more expressive than $\mathcal{L}^{QU}$? That depends on the underlying semantics.

In the case of probability, it is easy to see that it is not. Using additivity and affine homogeneity, it is easy to take an arbitrary formula $f \in \mathcal{L}^E$ and find a formula $f' \in \mathcal{L}^E$ that is equivalent to $f$ (with respect to structures in $\mathcal{M}^{prob}$) such that $e$ is applied only to propositional formulas. Then using the equivalence of $e(\varphi)$ and $\ell(\varphi)$, we can find a formula $f^T \in \mathcal{L}^{QU}$ equivalent to $f$ with respect to structures in $\mathcal{M}^{prob}$. It should be clear that the translation $f$ to $f^T$ causes at most a linear blowup in the size of the formula.

The same is true if we interpret formulas with respect to $\mathcal{M}^{bel}$ and $\mathcal{M}^{poss}$. In both cases, given a formula $f \in \mathcal{L}^E$, we can use (5) to find a formula $f' \in \mathcal{L}^E$ equivalent to $f$ such that $e$ is applied only to propositional formulas. (The details are in the full paper.) It is then easy to find a formula $f^T \in \mathcal{L}^{QU}$ equivalent to $f'$ with respect to structures in $\mathcal{M}^{bel}$ and $\mathcal{M}^{poss}$. However, now the translation from $f$ to $f^T$ can cause an exponential blowup in the size of the formula.

What about lower expectation/probability? In this case, $\mathcal{L}^E$ is strictly more expressive than $\mathcal{L}^{QU}$. It is not hard to construct two structures in $\mathcal{M}^{lp}$ that agree on all formulas in $\mathcal{L}^{QU}$ but disagree on formulas in $\mathcal{L}^E$ such as $e(p+q) > 1/2$. That means that there cannot be a formula in $\mathcal{L}^{QU}$ equivalent to $e(p+q) > 1/2$.

The following theorem summarizes this discussion.

**Theorem 4.1:** $\mathcal{L}^E$ and $\mathcal{L}^{QU}$ are equivalent in expressive power with respect to $\mathcal{M}^{prob}$, $\mathcal{M}^{bel}$, and $\mathcal{M}^{poss}$. $\mathcal{L}^E$ is strictly more expressive than $\mathcal{L}^{QU}$ with respect to $\mathcal{M}^{lp}$.

## 5 AXIOMATIZING EXPECTATION

In FHM, a sound and complete axiomatization is provided for $\mathcal{L}^{QU}$ both with respect to $\mathcal{M}^{prob}$ and $\mathcal{M}^{bel}$; in HP, a sound and complete axiomatization is provided for $\mathcal{L}^{QU}$ with respect to $\mathcal{M}^{lp}$. Here we provide a sound and complete axiomatization for $\mathcal{L}^E$ with respect to these structures.

The axiomatization for $\mathcal{L}^{QU}$ given in FHM splits into three parts, dealing respectively with propositional reasoning, reasoning about linear inequalities, and reasoning about likelihood. We follow the same pattern here. The following axioms characterize propositional reasoning:

**Taut**. All instances of propositional tautologies in the language $\mathcal{L}^E$.

**MP**. From $f$ and $f \Rightarrow g$ infer $g$.

Instances of **Taut** include all formulas of the form $f \vee \neg f$, where $f$ is an expectation formula. We could replace **Taut** by a simple collection of axioms that characterize propositional reasoning (see, for example, [Mendelson 1964]), but we have chosen to focus on aspects of reasoning about expectations.

The following axiom characterizes reasoning about linear inequalities:

**Ineq**. All instances in $\mathcal{L}^E$ of valid formulas about linear inequalities

This axiom is taken from FHM. There, an inequality formula is taken to be a Boolean combination of formulas of the form $a_1 x_1 + \cdots + a_n x_n \geq c$, over variables $x_1, \ldots, x_n$. Such a formula is valid if the resulting inequality holds under every possible assignment of real numbers to variables. To get an instance of **Ineq**, we replace each variable $x_i$ that occurs in a valid formula about linear inequalities by a primitive expectation term of the form $e(\gamma_i)$ (naturally each occurrence of the variable $x_i$ must be replaced by the same primitive expectation term $e(\gamma_i)$). As with **Taut**, we can replace **Ineq** by a sound and complete axiomatization for Boolean combinations of linear inequalities. One such axiomatization is given in FHM.

The following axioms characterize probabilistic expectation in terms of the properties described in Proposition 2.1.

**E1.** $e(\gamma_1 + \gamma_2) = e(\gamma_1) + e(\gamma_2)$,

**E2.** $e(a\varphi) = ae(\varphi)$ for all $a \in \mathbb{R}$,

**E3**. $e(\mathit{false}) = 0$,

**E4**. $e(\mathit{true}) = 1$,

**E5** $e(\gamma_1) \leq e(\gamma_2)$ if $\gamma_1 \leq \gamma_2$ is an instance of a valid formula about propositional gamble inequality (see below).

Axiom **E1** is simply additivity of expectations. Axioms **E2**, **E3**, and **E4**, in conjunction with additivity, capture affine homogeneity. Axiom **E5** captures monotonicity. A propositional gamble inequality is a formula of the form $\gamma_1 \leq \gamma_2$, where $\gamma_1$ and $\gamma_2$ are propositional gambles. Examples of valid propositional gamble inequalities are $p = p \wedge q + p \wedge \neg q$, $\varphi \leq \varphi + \psi$, and $\varphi \leq \varphi \vee \psi$. We define the semantics of gamble inequalities more carefully in Section 6, where we provide a complete axiomatization for them. As in the case of **Ineq**, we can replace **E5** by a sound and complete axiomatization for Boolean combinations of gamble inequalities.[5]

Let $\mathbf{AX}^{prob}$ be the axiomatization {**Taut, MP, Ineq, E1, E2, E3, E4, E5**}. As usual, given an axiom system **AX**, we say that a formula $f$ is *AX-provable* if it can be proved using the axioms and rules of inferences of **AX**. **AX** is sound with respect to a class $\mathcal{M}$ of structures if every **AX**-provable formula is valid in $\mathcal{M}$. **AX** is is *complete* with respect to $\mathcal{M}$ if every formula that is valid in $\mathcal{M}$ is **AX**-provable.

**Theorem 5.1:** $AX^{prob}$ *is a sound and complete axiomatization of* $\mathcal{L}^E$ *with respect to* $\mathcal{M}^{prob}$.

The characterizations of Theorems 2.4 and 2.9 suggest the appropriate axioms for reasoning about lower expectations and expected beliefs.

The following axioms capture the properties specified in Proposition 2.3:

**E6**. $e(\gamma_1 + \gamma_2) \geq e(\gamma_1) + e(\gamma_2)$,

**E7**. $e(a\gamma + b\, \mathit{true}) = ae(\gamma) + b$, for all $a, b \in \mathbb{R}$, $a \geq 0$,

**E8**. $e(a\gamma + b\, \mathit{false}) = ae(\gamma)$, for all $a, b \in \mathbb{R}$, $a \geq 0$.

Axiom **E6** is superadditivity of the expectation. Axioms **E7** and **E8** capture positive affine homogeneity. Note that because we do not have additivity, we cannot get away with simpler axioms as in the case of probability.

---

[5]We could have taken a more complex language that contains both expectation formulas and gamble inequalities. We could then merge the axiomatizations for expectation formulas and gamble inequalities. For simplicity, and to clarify the relationship between reasoning about expectation versus reasoning about likelihood (see Section 4), we consider only the restricted language in this paper.

Monotonicity is captured, as in the case of probability measures, by axiom **E5**. Let $\mathbf{AX}^{lp}$ be the axiomatization {**Taut, MP, Ineq, E5, E6, E7, E8**}.

**Theorem 5.2:** $AX^{lp}$ *is a sound and complete axiomatization of* $\mathcal{L}^E$ *with respect to* $\mathcal{M}^{lp}$.

Although it would seem that Theorem 5.2 should follow easily from Proposition 2.3, this is, unfortunately, not the case. As usual, soundness is straightforward, and to prove completeness, it suffices to show that if a formula $f$ is consistent with $\mathbf{AX}^{lp}$, it is satisfiable in a structure in $\mathcal{M}^{lp}$. Indeed, it suffices to consider formulas $f$ that are conjunctions of expectation inequalities and their negations. However, the usual approach for proving completeness in modal logic, which involves considering maximal consistent sets and canonical structures does not work. The problem is that there are maximal consistent sets of formulas that are not satisfiable. For example, there is a maximal consistent set of formulas that includes $e(p) > 0$ and $e(p) \leq 1/n$ for $n = 1, 2, \ldots$; this is clearly unsatisfiable. A similar problem arises in the completeness proofs for $\mathcal{L}^{QU}$ given in FHM and HP, but the techniques used there do not seem to suffice for dealing with expectations.

Of course, it is the case that any expectation function that satisfies the constraints in the formula $f$ and also every instance of axioms **E6**, **E7**, and **E8** must be a lower expectation, by Theorem 2.4. The problem is that, *a priori*, there are infinitely many relevant instances of the axioms. To get completeness, we must reduce this to a finite number of instances of these axioms. It turns out that this can be done, using techniques from linear programming and Walley's [1991] notion of *natural extension*. We leave the (quite nontrivial!) details to the full paper.

It is also worth noting that, although $\mathcal{L}^E$ is a more expressive language than $\mathcal{L}^{QU}$ in the case of lower probability/expectation, the axiomatization for $\mathcal{L}^E$ in this case is much more elegant than the corresponding axiomatization for $\mathcal{L}^{QU}$ given in HP.

We next consider expectation with respect to belief. As expected, the axioms capturing the interpretation of belief expectation rely on the properties pointed out in Proposition 2.8. Stating these properties in the logic requires a way to express the max and min of two propositional gambles. It turns out that we can view the notation $\gamma_1 \vee \gamma_2$ as an abbreviation for a more complex expression. Given a propositional gamble $\gamma = b_1\varphi_1 + \cdots + b_n\varphi_n$, we construct an equivalent gamble $\gamma'$ as follows. First define a family $\rho_A$ of propositional formulas indexed by $A \subseteq \{1, \ldots, n\}$ by taking $\rho_A = \bigwedge_{i \in A} \varphi_i \wedge (\bigwedge_{\{j \notin A\}} \neg \varphi_j)$. Thus, $\rho_A$ is true exactly if the $\varphi_i$'s for $i \in A$ are true, and the other $\varphi_j$'s are false. Note that the formulas $\rho_A$ are mutually exclusive. Define the real numbers $b_A$ for $A \subseteq \{1, \ldots, n\}$ by by taking $b_A = \sum_{i \in A} b_i$. Define $\gamma' = \sum_{A \subseteq \{1, \ldots, n\}} b_A \rho_A$.

It is easy to check that the propositional gambles $\gamma$ and $\gamma'$ are equal. Given two propositional gambles, say $\gamma_1$ and $\gamma_2$, we can assume without loss of generality that the involve the same primitive propositions $\varphi_1, \ldots, \varphi_n$. (If not, we can always add "dummy" terms of the form $0\psi$.) Form the gambles $\gamma_1'$ and $\gamma_2'$ as above. Since all the formulas mentioned in $\gamma_1'$ and $\gamma_2'$ are mutually exclusive, it it follows that $\max(\gamma_1', \gamma_2') = \sum_{A \subseteq \{1,\ldots,n\}} \max(b_A, b_A')\rho_A$. We take $\gamma_1 \vee \gamma_2$ to be an abbreviation for this gamble. (Note that if $\gamma_1$ and $\gamma_2$ are the propoisitional formulas $\varphi_1$ and $\varphi_2$, respectively, then, $\gamma_1 \vee \gamma_2$ really is a gamble equivalent to the propositional formula $\gamma_1 \vee \gamma_2$, so the use of $\vee$ is justified here.) Of course, we can similarly define $\gamma_1 \wedge \gamma_2$, simply by taking min instead of max.

With these definitions, the following axioms account for properties (4) and (5):

**E9.** $e(\gamma_1 \vee \cdots \vee \gamma_n) = \sum_{i=1}^{n} \sum_{\{I \subseteq \{1,\ldots,n\}:|I|=i\}} (-1)^{i+1} e(\bigwedge_{j \in I} \gamma_j)$.

**E10.** $e(b_1\varphi_1 + \cdots + b_n\varphi_n) = b_1 e(\varphi_1) + \cdots + b_n e(\varphi_n)$ if $\varphi_n \Rightarrow \varphi_{n-1}, \cdots, \varphi_3 \Rightarrow \varphi_2, \varphi_2 \Rightarrow \varphi_1$ are propositional tautologies.

Let $AX^{bel}$ be the axiomatization $\{\text{Taut}, \text{MP}, \text{Ineq}, \text{E5}, \text{E7}, \text{E8}, \text{E9}, \text{E10}\}$.

**Theorem 5.3:** $AX^{bel}$ is a sound and complete axiomatization of $\mathcal{L}^E$ with respect to $\mathcal{M}^{bel}$.

Finally, we consider expectation with respect to possibility. The axioms capturing the interpretation of possibilistic expectation $E_{\text{Poss}}$ rely on the properties pointed out in Proposition 2.11. The following axiom accounts for property (6):

**E11.** $(e(\varphi_1) \geq e(\varphi_2)) \Rightarrow (e(\varphi_1 \vee \varphi_2) = e(\varphi_1))$.

E11 is really a finitary version of (6); it essentially says that $e(\varphi_1 \vee \varphi_2) = \max(e(\varphi_1), e(\varphi_2))$. This finitary version turns out to suffice for completeness

Let $AX^{poss}$ be the axiomatization $\{\text{Taut}, \text{MP}, \text{Ineq}, \text{E5}, \text{E7}, \text{E8}, \text{E10}, \text{E11}\}$.

**Theorem 5.4:** $AX^{poss}$ is a sound and complete axiomatization of $\mathcal{L}^E$ with respect to $\mathcal{M}^{poss}$.

# 6 REASONING ABOUT GAMBLE INEQUALITIES

The axiomatization of Section 3 relied on an axiomatization of gamble inequalities. In this section, we provide such an axiomatization.

Let $\mathcal{L}^g$ consist of all Boolean combinations of gamble inequalities $\gamma_1 \geq \gamma_2$, where $\gamma_1$ and $\gamma_2$ are propositional gambles, as defined in Section 3. We use obvious abbreviations when needed, as in Section 3.

We can assign a semantics to gamble formulas by considering structure $M = (W, \pi)$ where $W$ is a set of worlds and $\pi$ associates with each world in $W$ a truth assignment on the primitive propositions. Let $\mathcal{M}^g$ be the class all such structures. (Clearly, every model in $\mathcal{M}^{prob}$, $\mathcal{M}^{lp}$, $\mathcal{M}^{bel}$, and $\mathcal{M}^{poss}$ can be interpreted as a model in $\mathcal{M}^g$, by simply "forgetting" the uncertainty measure over the worlds.) Then

$M \models \gamma_1 \geq \gamma_2$ iff
for all $w \in W$, $\{\!|\gamma_1|\!\}_M(w) \geq \{\!|\gamma_2|\!\}_M(w)$.

Again, Boolean combinations are given semantics in the obvious way.

We can axiomatize this logic as follows. As before, we have **Taut**, **MP**, and **Ineq** (although now we consider the instances of valid propositional tautologies and valid inequality formulas in the language $\mathcal{L}^g$). We then need only two more axioms to get a complete axiomatization. They correspond to the following two properties of indicator functions: for any $w \in W$, $X_U(w) + X_V(w) = X_{U \cup V}(w)$ if $U \cap V = \emptyset$, and $X_U(w) \leq X_V(w)$ if $U \subseteq V$.

**G1.** $\varphi \vee \psi = \varphi + \psi$ if $\varphi \wedge \psi \Leftrightarrow \mathit{false}$ is a propositional tautology,

**G2.** $\varphi \leq \psi$ if $\varphi \Rightarrow \psi$ is a propositional tautology.

Let $AX^g$ be the axiomatization $\{\text{Taut}, \text{MP}, \text{Ineq}, \text{G1}, \text{G2}\}$.

**Theorem 6.1:** $AX^g$ is a sound and complete axiomatization of $\mathcal{L}^g$ with respect to $\mathcal{M}^g$.

# 7 DECISION PROCEDURES

In FHM, it was shown that the satisfiability problem for $\mathcal{L}^{QU}$ was NP-complete, both with respect to $\mathcal{M}^{prob}$ and $\mathcal{M}^{bel}$; in HP, NP-completeness was also shown with respect to $\mathcal{M}^{lp}$. Here we prove similar results for the language $\mathcal{L}^E$. In the case of $\mathcal{M}^{prob}$, this is not at all surprising, given Theorem 4.1 and the fact that the translation from $\mathcal{L}^E$ to $\mathcal{L}^{QU}$ causes only a linear blowup in the case of $\mathcal{M}^{prob}$. However, we cannot get the result for $\mathcal{M}^{bel}$ or $\mathcal{M}^{poss}$ from Theorem 4.1, since the translation causes an exponential blowup. Of course, in the case of $\mathcal{M}^{lp}$, no translation exists at all. Nevertheless, in all these cases, we can get NP-completeness using techniques very much in the spirit of the linear programming techniques used in FHM.

Note that if we allow real numbers as coefficients in expectation formulas, we have to carefully discuss the issue of representation of such numbers. To avoid these complications, we restrict $\mathcal{L}^E$ in this section to allow only integer coefficients. (Similar restrictions are also made in FHM and HP when dealing with complexity issues.) Let $\mathcal{L}_1^E$ be the resulting language. We can still express rational coefficients in $\mathcal{L}_1^E$ by the standard trick of "clearing the denominator".

**Theorem 7.1:** *The problem of deciding whether a formula in $\mathcal{L}_1^E$ is satisfiable in $\mathcal{M}^{prob}$ (resp., $\mathcal{M}^{lp}$, $\mathcal{M}^{bel}$, $\mathcal{M}^{poss}$) is NP-complete.*

Perhaps not surprisingly, the logic for reasoning about gamble inequalities discussed in Section 6 is also NP-complete.

**Theorem 7.2:** *The problem of deciding whether a formula of $\mathcal{L}^g$ is satisfiable in $\mathcal{M}^g$ is NP-complete.*

### Acknowledgments

We thank the UAI referees for their comments. This work was supported in part by NSF under grant IIS-0090145, by ONR under grants N00014-00-1-0341, N00014-01-1-0511, and N00014-02-1-0455, by the DoD Multidisciplinary University Research Initiative (MURI) program administered by the ONR under grants N00014-97-0505 and N00014-01-1-0795. In addition, Halpern was supported by a Guggenheim and a Fulbright Fellowship. Sabbatical support from CWI and the Hebrew University of Jerusalem is also gratefully acknowledged.

### References

Borel, E. (1943). *Les Probabilités et la Vie*. Paris: Presses Universitaires de France. English translation *Probabilities and Life* (1962), New York: Dover.

Choquet, G. (1953). Theory of capacities. *Annales de l'Institut Fourier (Grenoble) 5*, 131–295.

Dempster, A. P. (1967). Upper and lower probabilities induced by a multivalued mapping. *Annals of Mathematical Statistics 38*, 325–339.

Dubois, D. and H. Prade (1982). On several representations of an uncertain body of evidence. In M. M. Gupta and E. Sanchez (Eds.), *Fuzzy Information and Decision Processes*, pp. 167–181.

Dubois, D. and H. Prade (1987). The mean value of a fuzzy number. *Fuzzy Sets and Systems 24*, 297–300.

Dubois, D. and H. Prade (1990). An introduction to possibilistic and fuzzy logics. In G. Shafer and J. Pearl (Eds.), *Readings in Uncertain Reasoning*, pp. 742–761. San Francisco, Calif.: Morgan Kaufmann.

Fagin, R., J. Y. Halpern, and N. Megiddo (1990). A logic for reasoning about probabilities. *Information and Computation 87*(1/2), 78–128.

Halpern, J. Y. (2002). Reasoning about uncertainty. Book manuscript.

Halpern, J. Y. and R. Pucella (2001). A logic for reasoning about upper probabilities. In *Proceedings of the Seventeenth Conference on Uncertainty in Artificial Intelligence*, pp. 203–210.

Huber, P. J. (1981). *Robust Statistics*. Wiley Series in Probability and Mathematical Statistics. Wiley Interscience.

Mendelson, E. (1964). *Introduction to Mathematical Logic*. New York: Van Nostrand.

Schmeidler, D. (1989). Subjective probability and expected utility without additivity. *Econometrica 57*, 571–587.

Shafer, G. (1976). *A Mathematical Theory of Evidence*. Princeton, N.J.: Princeton University Press.

Smith, C. A. B. (1961). Consistency in statistical inference and decision. *Journal of the Royal Statistical Society, Series B 23*, 1–25.

Walley, P. (1981). Coherent lower (and upper) probabilities. Manuscript, Dept. of Statistics, University of Warwick.

Walley, P. (1991). *Statistical Reasoning with Imprecise Probabilities*. Chapman and Hall.

Wilson, N. and S. Moral (1994). A logical view of probability. In A. G. Cohn (Ed.), *Proceedings of the 11th European Conference on Artificial Intelligence (ECAI-94)*, pp. 71–95. John Wiley.